\documentclass{article}

\usepackage{times}
\usepackage{graphicx} 
\usepackage{subfigure} 
\usepackage{multirow}
\usepackage{makecell}
\usepackage{courier}

\usepackage{natbib}

\usepackage{algorithm}
\usepackage{algorithmic}
\usepackage{ctable} 

\usepackage{float}

\usepackage{hyperref}

\usepackage[accepted]{icml2017}

\icmltitlerunning{Benchmark Environments for Multitask Learning in Continuous Domains}

\begin{document} 

\twocolumn[
\icmltitle{Benchmark Environments for Multitask Learning in Continuous Domains}



\icmlsetsymbol{equal}{*}

\begin{icmlauthorlist}
\icmlauthor{Peter Henderson}{mcgill} 
\icmlauthor{Wei-Di Chang}{mcgill}
\icmlauthor{Florian Shkurti}{mcgill}
\icmlauthor{Johanna Hansen}{mcgill}
\icmlauthor{David Meger}{mcgill}
\icmlauthor{Gregory Dudek}{mcgill}
\end{icmlauthorlist}

\icmlaffiliation{mcgill}{McGill University, Montreal, Quebec, Canada}

\icmlcorrespondingauthor{Peter Henderson}{peter.henderson@mail.mcgill.ca}

\icmlkeywords{multi-task learning, transfer learning, reinforcement learning, continuous domains}

\vskip 0.3in
]



\printAffiliationsAndNotice{}  

\begin{abstract} 
As demand drives systems to generalize to various domains and problems, the study of multitask, transfer and lifelong learning has become an increasingly important pursuit.  In discrete domains, performance on the Atari game suite has emerged as the \textit{de facto} benchmark for assessing multitask learning. However, in continuous domains there is a lack of agreement on standard multitask evaluation environments which makes it difficult to compare different approaches fairly. In this work, we describe a benchmark set of tasks that we have developed in an extendable framework based on OpenAI Gym. We run a simple baseline using Trust Region Policy Optimization and release the framework publicly to be expanded and used for the systematic comparison of multitask, transfer, and lifelong learning in continuous domains.
\end{abstract}

\section{Introduction}
\label{intro}


Multitask learning involves training a single agent in a lifelong context across a series of tasks. When tasks share similar characteristics, there is potential for the learning agent to  reuse information, potentially achieving greater performance or learning more rapidly on down-stream tasks than would be possible by learning each task from scratch. 

The promise of multitask learning has been previously demonstrated in several contexts. It has been shown that transfer learning on multiple related tasks improves the ability of the agent to learn a larger variety of domains while using less training data overall~\cite{Thrun96NIPS}. This naturally renders the agent more applicable to real-world scenarios. Additionally, by training on multiple tasks, the agent can exploit common traits to gain efficiency and generalize to unseen tasks~\cite{MTLCaruana1997,adaptive2016,finn2016generalizing}.

\begin{figure}[t]
\hspace*{-.5cm}\includegraphics[width=0.21\textwidth]{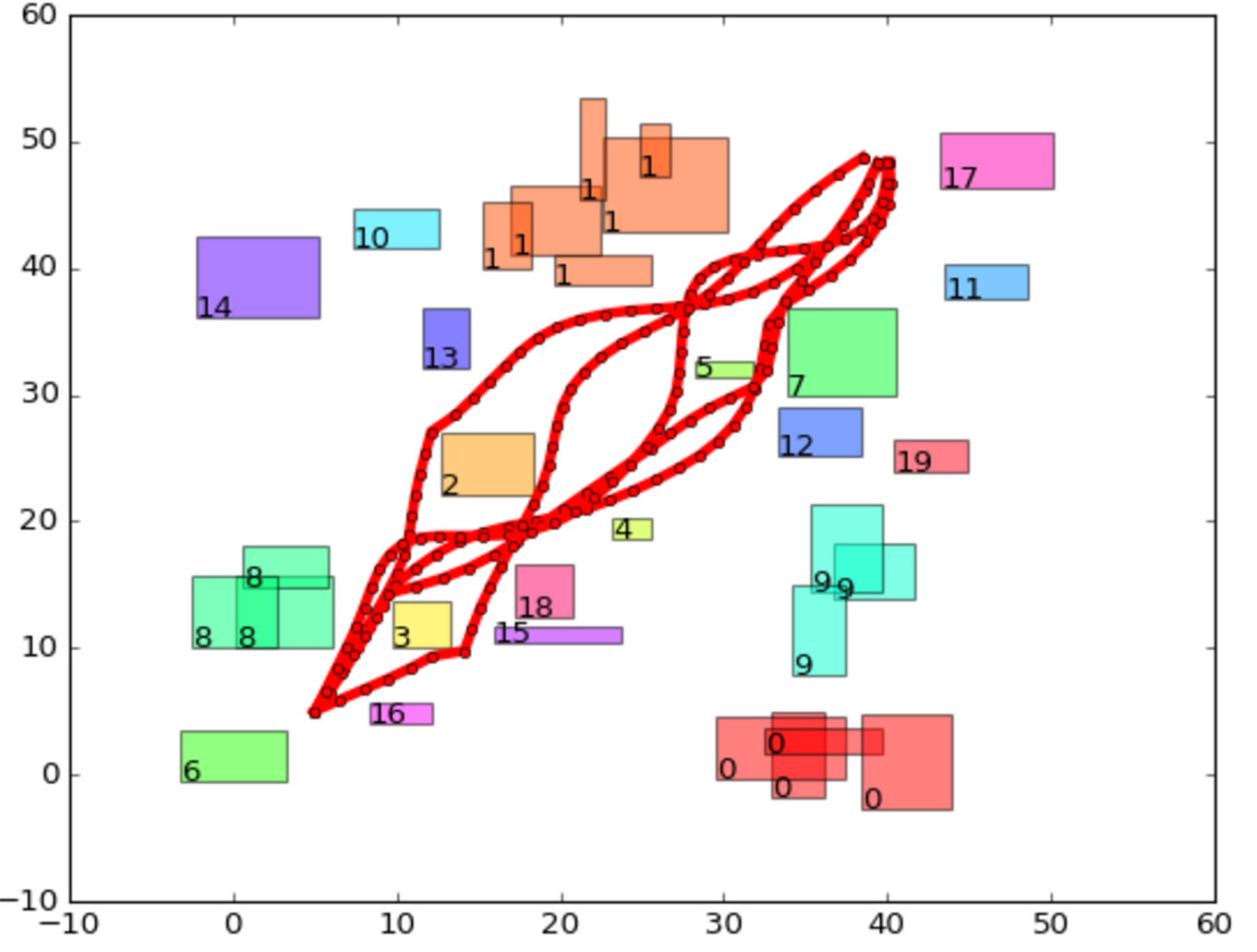}
\raisebox{0.1\height}{\includegraphics[width=0.17\textwidth]{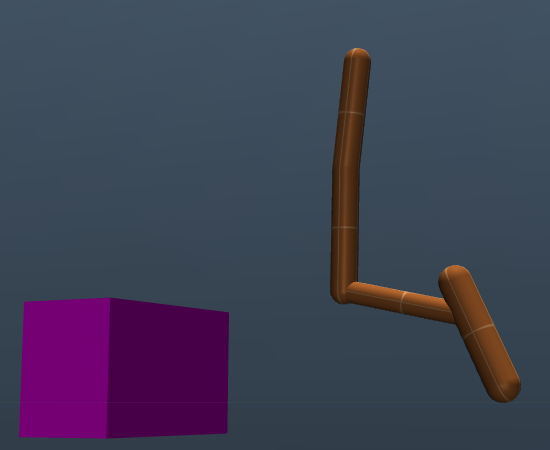}}
\centering\includegraphics[width=0.14\textwidth]{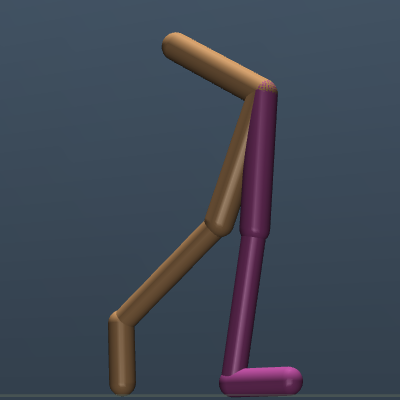}
\centering\includegraphics[width=0.14\textwidth]{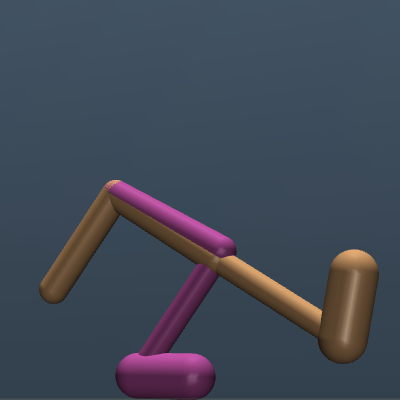}
\caption{Example environments: 2D navigation task, with several sample paths (top-left). Hopper with a wall (top-right). Walker2d with big (bottom-right) and small (bottom-left) feet.\\}
\vspace{-2em}
\end{figure}


In discrete domains, several works have investigated the transferring of playing knowledge acquired between various Atari games. It is intuitive that there should be some knowledge transfer between Atari games (Breakout and Pong are similar in catching a ball; DemonAttack, Carnaval, Assault, and AirRaid share a goal in shooting upwards to destroy enemies). However, despite several authors' attempts to demonstrate multitask learning in both Atari tasks and the DeepMind Lab Labyrinth~\cite{parisotto2015actor, rusu2015policy, jaderberg2016reinforcement}, this task that has yet to be convincingly solved. Atari video-games therefore stand as a useful benchmark which can be used to induce further progress in multitask, transfer and lifelong learning in discrete domains. On the contrary, in continuous domains, each approach for continuous domain multitask learning has utilized a unique set of environments with little mutual overlap. As such, there is a need for such open-source standard benchmarks.

This problem is one that has been recognized by other research groups. Particularly, OpenAI lists the need for benchmark environments and investigation of multitask learning in continuous domains in a request for research\footnote{See: \scriptsize\url{https://openai.com/requests-for-research/\#multitask-rl-with-continuous-actions}}. As they mention, the current OpenAI Gym~\cite{gym2016} environments do not share enough characteristics to likely pose as effective multitask or lifelong learning benchmarks. The contribution of this work is a set of benchmark environments that are suitable to evaluate continuous domain multitask learning. Our environments are constructed using an expandable software framework built on top of OpenAI Gym. Here, we show over 50 new environment variations (spread among 12 broad groups of variation types) for challenging new continuous domain tasks. We verify the utility of these environments for evaluating multitask learning by reporting the performance of a well-known reinforcement learning algorithm on our multitask benchmark environments as a simple baseline.

\section{Related Work}

Several works investigate multitask or transfer learning with MuJoCo tasks. These tasks include: navigating around a wall (where a wall separates an agent from its goal); the OpenAI Gym Reacher environment with an added image state space of the environment; jumping over a wall using a model similar to the OpenAI Half-Cheetah environment~\cite{finn2016generalizing}; varying the gravity of various standard OpenAI Gym benchmark environments (Reacher, Hopper, Humanoid, HalfCheetah) and transferring between the modified environments; adding motor noise to the same set of environments~\cite{christiano2016transfer}; simulated grasping and stacking using a Jaco arm~\cite{rusu2016sim}; and several custom grasping and manipulation tasks to demonstrate learning invariant feature spaces~\cite{gupta2017learning}.

Other works investigate using classical control systems and robotics simulations with a set of varied hyperparameters for each environment. These include: a simple mass spring damper task, cart-pole with continuous control; a three-link inverted pendulum with continuous control; a quadrotor control task~\cite{ammar2014online}; a double-linked pendulum task; a modified cartpole balancing task which can transfer to physical system~\cite{higueraadapting}.

\section{Environments} 
In our initial release of the \textit{gym-extensions} framework\footnote{Found at: {\scriptsize \url{https://github.com/Breakend/gym-extensions/}}. Pull requests and issues are welcome. More details for each environment will be provided in the open-source repository as well as a place to upload new benchmark algorithm runs.}, we include a number of modifications of the standard gym environments as well as novel continuous domains, and provide a framework which allows easy modification of environment characteristics.

\begin{table*}[!htbp]
\centering\scriptsize\begin{tabular}{| l | c | c | c | c | c |}
\hline
\textbf{Environment} & \textbf{Fully Trained} & \textbf{After Env Training} & \textbf{First Step} & \textbf{Single Env} & \textbf{Grouping Description}\\
\specialrule{.1em}{.05em}{.05em} 
HopperGravityHalf-v0 & 1495.93 $\pm$ 823.51 & 2352.19 $\pm$ 580.53 & 13.48 $\pm$ 8.71& 1843.89 $\pm$ 485.25 & \multirow{5}{*}{\makecell{Environment with \\different gravity \\parameters\\(as specified in the name)}}\\
HopperGravityThreeQuarters-v0 & 413.77 $\pm$ 252.67 & 2245.13 $\pm$ 872.16  & 697.96 $\pm$ 210.79& 2328.09 $\pm$ 834.35&\\
Hopper-v1 & 668.52 $\pm$ 159.90 &  2622.31 $\pm$ 1032.45 & 781.88 $\pm$ 262.35 & 3232.87 $\pm$ 582.55&\\
HopperGravityOneAndQuarter-v0 & 922.76 $\pm$ 128.71 & 3006.17 $\pm$ 847.30 & 818.08 $\pm$ 255.85 & 3028.04 $\pm$ 875.81 &\\
HopperGravityOneAndHalf-v0 & 2690.57 $\pm$ 1110.39 & 2792.72 $\pm$ 1075.30 & 658.15 $\pm$ 117.14 & 2169.07 $\pm$ 825.75 &\\
\hline
Total for Grouping & 990.95 $\pm$ 1022.32 & 2603.704 $\pm$ 881.54  &593.91 $\pm$ 184.43&2520.39 $\pm$ 720.74&\\
\specialrule{.1em}{.05em}{.05em} 
Walker2dGravityHalf-v0 & 1366.07 $\pm$ 1126.59   & 3485.19 $\pm$ 1054.06& 5.35 $\pm$ 10.30 & 2231.86 $\pm$ 902.31& \multirow{5}{*}{\makecell{Same as above}}\\
Walker2dGravityThreeQuarters-v0 & 3686.37 $\pm$ 287.96  & 3962.69 $\pm$ 1061.71&1071.95 $\pm$ 267.35 & 2431.87 $\pm$ 935.14 &\\
Walker2d-v1 & 4030.00 $\pm$ 85.76  & 3732.04 $\pm$ 1314.89 & 930.92 $\pm$ 264.88& 2570.15 $\pm$ 915.58&\\
Walker2dGravityOneAndQuarter-v0 & 4115.23 $\pm$ 90.33  & 4090.30 $\pm$ 1058.62 & 926.06 $\pm$ 303.76& 3505.52 $\pm$ 1626.58&\\
Walker2dGravityOneAndHalf-v0 & 4201.08 $\pm$ 684.37  & 3988.62 $\pm$ 971.43 & 925.93 $\pm$ 290.33 & 2435.21 $\pm$ 1391.00&\\
\hline
Total for Grouping & 3479.76 $\pm$ 1230.72  & 3851.768 $\pm$ 1092.1 &772.04 $\pm$ 227.32& 2634.92 $\pm$ 1154.12 &\\
\specialrule{.1em}{.05em}{.05em} 
HalfCheetahGravityHalf-v0 & 1495.93 $\pm$ 823.51 & 1107.50 $\pm$ 784.31 & -369.31 $\pm$ 113.71 & 2048.93 $\pm$ 761.03& \multirow{5}{*}{\makecell{Same as above}}\\
HalfCheetahGravityThreeQuarters-v0 & 1671.76 $\pm$ 594.15 & 2142.78 $\pm$ 818.99 & 1410.25 $\pm$ 529.41&3268.26 $\pm$ 703.43&\\
HalfCheetah-v1 &  1743.97 $\pm$ 100.32 &  2410.50 $\pm$ 137.30 &1867.99 $\pm$ 251.58 & 2554.01 $\pm$ 115.69 &\\
HalfCheetahGravityOneAndQuarter-v0 & 2649.13 $\pm$ 143.43 & 2939.14 $\pm$ 164.62 & 1966.66$\pm$ 171.88 & 2572.64 $\pm$ 90.80&\\
HalfCheetahGravityOneAndHalf-v0 & 3421.21 $\pm$ 165.60 & 3402.83 $\pm$ 204.00 & 2143.76 $\pm$ 236.60& 2276.82 $\pm$ 93.30&\\
\hline
Total for Grouping & 2196.41 $\pm$ 867.75  & 2400.55 $\pm$ 421.84 &1403.87 $\pm$ 260.63&2544.13 $\pm$ 352.85&\\
\specialrule{.1em}{.05em}{.05em} 
HumanoidGravityHalf-v0 & 416.41 $\pm$ 76.41 & 421.12 $\pm$ 95.61 & 89.60 $\pm$ 10.92 & 849.29 $\pm$ 213.81 & \multirow{5}{*}{\makecell{Same as above}}\\
HumanoidGravityThreeQuarters-v0  & 356.74 $\pm$ 52.52 & 385.54 $\pm$ 72.98 & 293.14 $\pm$ 66.12 & 637.33 $\pm$ 170.51 &\\
Humanoid-v1 & 310.09 $\pm$ 55.31 & 326.59 $\pm$ 59.78 & 267.11 $\pm$ 52.74& 483.35 $\pm$ 106.12 &  \\
HumanoidGravityOneAndQuarter-v0 & 261.01 $\pm$ 31.75 & 269.03 $\pm$ 40.59 & 233.82 $\pm$ 39.41 & 576.98 $\pm$ 124.25 &\\
HumanoidGravityOneAndHalf-v0 & 227.17 $\pm$ 33.62 & 226.94 $\pm$ 29.09 & 208.74 $\pm$ 34.43 & 538.24 $\pm$ 113.17 & \\
\hline
Total for Grouping & 314.28 $\pm$ 85.41 &325.84 $\pm$ 59.61 &218.48 $\pm$ 40.72 & 617.03 $\pm$ 145.57 &\\
\specialrule{.1em}{.05em}{.05em} 

\end{tabular}
\caption{{\small Average and standard deviation ($\mu \pm \sigma$) of reward across a set of 20 sample rollouts. We show samples immediately after training on a particular environment and the reward obtained by the final trained policy on all previously seen environment. A group of tasks is defined by a bold separator and the total average across all final rollouts is presented. ``Fully Trained'' lists the final evaluation result using the fully trained policy which has seen all the environments. ``After Env Training'' lists the evaluation immediately after training on that specific environment (having seen all the previous environments up until that point in the group). The ``First Step'' column indicates the reward at the first iteration of training on the new environment after having trained on the previous environments in the group. ``Single Env'' indicates rollouts on a policy trained solely on that environment (with all the same training parameters).}}
\label{table:results_gravity}
\end{table*}

\begin{table*}[!htbp]
\centering\scriptsize\begin{tabular}{| l | c | c | c | c |}
\hline
\textbf{Environment \footnote{In order of training}} & \textbf{Fully Trained} & \textbf{After Env Training} & \textbf{First Step} & \textbf{Grouping Description}\\
\specialrule{.1em}{.05em}{.05em} 
HopperSmallFoot-v0 &  591.91 $\pm$ 150.73 & 1330.65 $\pm$ 402.07 & 9.83 $\pm$ 4.52 & \multirow{8}{*}{\makecell{Environments with \\body part size\\ variations}}\\
HopperSmallLeg-v0 & 2074.58 $\pm$ 800.61 &  1359.85 $\pm$ 311.91 & 744.35  $\pm$120.50  &\\
HopperSmallThigh-v0 & 919.87 $\pm$ 343.57  &  1492.44 $\pm$ 486.72 & 1719.34 $\pm$757.42 &\\
HopperSmallTorso-v0 & 1094.85 $\pm$ 319.94 &  1492.97 $\pm$ 518.47 & 1636.30 $\pm$298.22  &\\
HopperBigFoot-v0 & 2823.58 $\pm$ 887.25 &  1819.91 $\pm$ 812.33 & 559.61 $\pm$145.24  &\\
HopperBigLeg-v0 & 1020.13 $\pm$ 454.74 &  2148.57 $\pm$ 795.95& 689.58 $\pm$96.23 &\\
HopperBigThigh-v0 & 2799.39 $\pm$ 748.89 &  1827.48 $\pm$ 767.09 & 674.14 $\pm$ 101.72 &\\
HopperBigTorso-v0 & 1971.50 $\pm$ 794.24 &  2090.68 $\pm$ 693.34 & 1110.46 $\pm$ 213.74 &\\
\hline
Total for Grouping & 1661.98 $\pm$ 1025.19 & 1695.31 $\pm$ 598.48 & 892.95 $\pm$ 203.69 &\\
\specialrule{.1em}{.05em}{.05em} 
Walker2dSmallFoot-v0 & 2497.10 $\pm$ 1309.80  & 531.08 $\pm$ 329.00 & -2.87 $\pm$ 2.56 &\multirow{8}{*}{\makecell{Same as above}}\\
Walker2dSmallLeg-v0 & 3181.14 $\pm$ 1131.29  & 1120.19 $\pm$ 597.09 & 318.20 $\pm$ 229.00 & \\
Walker2dSmallThigh-v0 & 3106.65 $\pm$ 641.34  & 1735.39 $\pm$ 880.44 & 1317.72 $\pm$ 737.66 &\\
Walker2dSmallTorso-v0 & 3132.88 $\pm$ 991.48  & 1838.79 $\pm$ 965.60 & 979.32 $\pm$ 582.14 & \\
Walker2dBigFoot-v0 & 2751.34 $\pm$ 1216.07 &  1873.60 $\pm$ 1047.41 & 789.32 $\pm$ 289.65 &\\
Walker2dBigLeg-v0 & 2820.94 $\pm$ 1108.26 & 2133.64 $\pm$ 1246.53 & 1743.53 $\pm$ 1106.90 & \\
Walker2dBigThigh-v0 & 892.54 $\pm$ 212.46 & 2756.79 $\pm$ 1238.62 & 805.31 $\pm $147.40 & \\
Walker2dBigTorso-v0 & 3097.45 $\pm$ 1383.43 & 2701.94 $\pm$ 1473.47 & 3045.06 $\pm$ 967.83 & \\
\hline
Total for Grouping & 2685.01 $\pm$ 1280.70  &  1836.42 $\pm$ 972.27 & 1124.45 $\pm$ 507.89 & \\
\specialrule{.1em}{.05em}{.05em} 
HalfCheetahSmallFoot-v0 & 2003.46 $\pm$ 933.59 & 898.51 $\pm$ 363.85 & -502.264 $\pm$ 97.99  &\multirow{8}{*}{\makecell{Same as above}}\\
HalfCheetahSmallLeg-v0 & 2327.16 $\pm$ 702.69 &  1494.03 $\pm$ 310.11 & 904.82 $\pm$ 330.37 &\\
HalfCheetahSmallThigh-v0 & 2555.16 $\pm$ 96.80  &  1672.22 $\pm$ 110.11 & 1311.60 $\pm$ 281.24 &\\
HalfCheetahSmallTorso-v0 & 2294.72 $\pm$ 109.20 & 1845.20 $\pm$ 86.03 & 1515.53 $\pm$ 94.68 & \\
HalfCheetahBigFoot-v0 & 2211.92 $\pm$ 65.81 & 1997.73 $\pm$ 101.36 & 1789.90 $\pm$ 82.31 &\\
HalfCheetahBigLeg-v0 & 2269.78 $\pm$ 95.57 & 2101.74 $\pm$ 95.98 & 1908.53 $\pm$ 99.49 &\\
HalfCheetahBigThigh-v0 & 2424.95 $\pm$ 94.19 &   2345.88 $\pm$ 381.33 & 1925.04 $\pm$ 347.83 & \\
HalfCheetahBigTorso-v0 & 2686.13 $\pm$ 97.96 & 2620.46 $\pm$ 297.88 & 2456.04 $\pm$ 421.03 & \\
\hline
Total for Grouping & 2346.66 $\pm$ 464.81 & 1871.97 $\pm$ 218.33& 1413.64 $\pm$ 207.11 &\\
\specialrule{.1em}{.05em}{.05em} 
HumanoidSmallFoot-v0 & 391.70 $\pm$ 124.75 & 228.46 $\pm$ 62.80 & 94.85  $\pm$ 106.57 & \multirow{13}{*}{\makecell{Same as above}}\\
HumanoidSmallLeg-v0 & 438.90 $\pm$ 113.80 & 290.88 $\pm$ 82.86  & 253.81 $\pm$ 68.12 & \\
HumanoidSmallThigh-v0 & 378.47 $\pm$ 113.70 & 347.38 $\pm$ 99.89  & 322.97 $\pm$ 93.28 & \\
HumanoidSmallTorso-v0 & 433.04 $\pm$ 88.76 & 341.22 $\pm$ 89.69 & 313.71 $\pm$ 82.52 & \\
HumanoidBigFoot-v0 & 456.39 $\pm$ 85.60 & 399.96 $\pm$ 95.84  & 355.16 $\pm$ 92.07 & \\
HumanoidBigLeg-v0 & 430.82 $\pm$ 105.41  & 380.20 $\pm$ 97.58  & 347.26 $\pm$ 87.78 & \\
HumanoidBigThigh-v0 & 365.79 $\pm$ 72.06 & 331.80 $\pm$ 84.06  & 303.13 $\pm$ 77.17 & \\
HumanoidBigTorso-v0 & 397.91 $\pm$ 109.04 & 392.66 $\pm$ 108.20 & 374.94 $\pm$ 102.12 &  \\
HumanoidSmallHead-v0 & 422.33 $\pm$ 112.20  & 395.75 $\pm$ 101.70  & 386.14 $\pm$ 96.37 & \\
HumanoidBigHead-v0 & 507.29 $\pm$ 146.50  & 409.66 $\pm$ 109.13  & 411.98 $\pm$ 119.99 & \\
HumanoidSmallArm-v0 & 429.93 $\pm$ 113.26  & 416.41 $\pm$ 94.45   & 400.16 $\pm$ 91.75 & \\
HumanoidBigArm-v0 & 466.13 $\pm$ 129.87   & 411.23 $\pm$ 111.20  & 392.53 $\pm$ 115.21 & \\
HumanoidSmallHand-v0 & 450.07 $\pm$  76.72  & 423.29 $\pm$ 101.61 & 417.45 $\pm$ 99.08 & \\
HumanoidBigHand-v0 & 409.46 $\pm$ 69.38  & 420.65 $\pm$ 100.29 & 415.04 $\pm$ 108.38 &  \\
\hline
Total for Grouping & 427.02 $\pm$  112.56 &  370.68 $\pm$ 95.66 & 342.08 $\pm$ 95.23 &\\
\specialrule{.1em}{.05em}{.05em} 

\end{tabular}
\caption{Results for modified body-part running task groups. Same parameters as described in Table~\ref{table:results_gravity}. Number of training iterations lowered to 500 per environment due to the larger number of environments.}
\label{table:results_body_parts}
\end{table*}

\begin{table*}[!htbp]
\centering\scriptsize\begin{tabular}{| l | c | c | c | c |}
\hline
\textbf{Environment \footnote{In order of training}} & \textbf{Fully Trained} & \textbf{After Env Training}& \textbf{First Step} & \textbf{Grouping Description}\\
\hline
HopperWithSensor-v0 & 747.67 $\pm$ 27.06 & 2881.79 $\pm$ 623.11 & 15.51 $\pm$ 14.88 & \multirow{2}{*}{\makecell{Envs with sensor readouts (reading zero if no wall)\\then permutated with a random wall in the runner path.}} \\
HopperWall-v0 & 687.58  $\pm$ 58.81 & 695.00 $\pm$ 93.70 & 695.94 $\pm$ 102.96 & \\
\hline
Walker2dWithSensor-v0 & 1897.74 $\pm$ 1101.13 & 3357.76 $\pm$ 1142.85 & -2.27 $\pm$ 8.84 &\multirow{2}{*}{\makecell{Same as above.}}\\
Walker2dWall-v0 & 1271.78 $\pm$ 881.57 & 974.83 $\pm$ 664.29 & 635.45 $\pm$ 303.73 &\\
\hline
HalfCheetahWithSensor-v0 & 2924.83 $\pm$ 165.69 & 2754.58 $\pm$ 151.81 & -296.32 $\pm$ 110.45 &\multirow{2}{*}{\makecell{Same as above.}}\\
HalfCheetahWall-v0 & 2022.90 $\pm$ 826.91 & 2159.17 $\pm$ 805.27 & 2043.85 $\pm$ 807.09 & \\
\hline
HumanoidWithSensor-v0 & 339.37 $\pm$ 47.38 & 285.70 $\pm$ 51.99 & 67.37 $\pm$ 8.31 & \multirow{2}{*}{\makecell{Same as above.}}\\
HumanoidWall-v0 & 334.03 $\pm$ 51.49 & 328.90 $\pm$ 57.41 & 284.55 $\pm$ 49.07 & \\
\specialrule{.1em}{.05em}{.05em} 
Humanoid-v1 &  252.38  $\pm$ 12.05 & 269.23 $\pm$ 75.11 & 72.398 $\pm$ 2.56& \multirow{3}{*}{\makecell{Learning to run, then standup, and finally do both\\ in one environment. }}\\
HumanoidStandup-v0 & 75861.96 $\pm$ 19951.32 & 75906.45 $\pm$ 22390.07 & 70659.6 $\pm$ 19479.4& \\
HumanoidStandupAndRun-v0 & 71269.19 $\pm$ 16689.99 & 73919.85 $\pm$ 19215.23 & 70021.9$\pm$18660.3&\\
\hline
HumanoidWithSensor-v0 & 112.74 $\pm$ 23.09 & 114.75 $\pm$ 13.48 & 64.6059 $\pm$ 1.76& \multirow{4}{*}{\makecell{Learning to run, then standup, and\\finally jump over a wall with sensor readouts\\if there is a wall within the ray-tracing max distance.}}\\
HumanoidStandupWithSensor-v0 & 53124.35 $\pm$ 15136.53 & 58335.81 $\pm$ 16259.60 & 52029.5 $\pm$ 13585.10  &  \\
HumanoidStandupAndRunWithSensor-v0 & 59263.15 $\pm$ 12285.51 & 62570.26 $\pm$ 14258.35 & 55929.7 $\pm$ 15432.20 & \\
HumanoidStandupAndRunWall-v0 &  61468.03 $\pm$ 16135.02 & 66789.60 $\pm$ 14405.80 & 61764.5 $\pm$ 15150.20 &  \\
\specialrule{.1em}{.05em}{.05em} 
\end{tabular}
\caption{Results for modified running tasks with sensors, walls, or multiple goals.}
\label{table:results_sensor}
\end{table*}

\begin{table*}[!htbp]
\centering\scriptsize\begin{tabular}{| l | c | c | c | c |}
\hline
\textbf{Environment \footnote{In order of training}} & \textbf{Fully Trained} & \textbf{After Env Training} & \textbf{First Step} & \textbf{Grouping Description}\\
\specialrule{.1em}{.05em}{.05em} 
Striker-v0 &  -124.87 $\pm$ 47.33 & -114.61 $\pm$ 36.93 & -590.61 $\pm$ 78.77& \multirow{2}{*}{\makecell{The OpenAI Gym Striker env and the same task\\with both random object start and goal position}}\\
StrikerMovingStart-v0 &  -163.08 $\pm$ 76.29 & -146.06 $\pm$ 60.21& -171.06 $\pm$ 92.10 &\\
\hline
Total for Grouping &  -143.97 $\pm$ 66.29 & -130.33 $\pm$ 43.57& -380.83 $\pm$ 85.44& \\
\specialrule{.1em}{.05em}{.05em} 
Pusher-v0 &  -24.83 $\pm$ 2.39  & -24.59 $\pm$ 4.01&-209.57 $\pm$ 7.46& \multirow{2}{*}{\makecell{The OpenAI Gym Pusher env and the same task\\with both random object start and goal position}}\\
PusherMovingGoal-v0 & -28.01 $\pm$ 7.24 & -27.76 $\pm$ 6.20& -34.90 $\pm$ 9.38& \\
\hline
Total for Grouping & -26.42 $\pm$ 5.62  &-26.17 $\pm$ 5.11&-122.24 $\pm$ 8.42& \\
\specialrule{.1em}{.05em}{.05em} 

\end{tabular}
\caption{Results for arm-based task groups. Same parameters as described in Table~\ref{table:results_gravity}.}
\label{table:results_arm}
\end{table*}

\subsection{Mujoco}
We base our modified environments on the existing ``running'' (Humanoid, Hopper, Half-Cheetah, and Walker2d) and ``arm-based'' (Pusher and Striker) environments in OpenAI Gym. First, we provide a high-level overview of our modifications and suggested grouping, then we show the specific environment names in our benchmarking results.

\subsubsection{Gravity modifications}
For the running agents, we provide ready environments with various scales of simulated earth-like gravity, ranging from one half to one and a half of the normal gravity level ($-4.91$ to $-12.26\,m\cdot s^{-2}$ in increments of $.25{g}_{{earth}}$). We propose that a successful multitask learning algorithm will extract the underlying walking action structure and reuse the applicable knowledge without forgetting how to walk in varying gravity conditions.


\subsubsection{Wall and Sensor environments} \label{sensor_envs}
Inspired by the wall jumping experiment in \cite{finn2016generalizing}, we build a set of similar environments by extending the OpenAI running tasks to use a multi-beam noiseless range sensor. We emit ray-beams from the torso of the runner for the measurements (with an arc of 90 degrees, 10 beams, a maximum sensing distance of 10 meters, and readouts normalized to a range of $[0,1]$). We provide the usual running tasks with the sensor perception enabled (with no readings since there is no wall), and extra environments with a wall set in the path of the agent at a location drawn from a uniform distribution from 1.8 to 3.8 meters ahead of the agent's start location.

\subsubsection{Morphology modifications}
For the running agents, we provide environments which vary the morphology of a specific body part of the agent. The modifications made to each agent are seen in Table~\ref{table:results_body_parts}. We define ``Big'' bodyparts as scaling the mass and width of the limb by $1.25$ and ``Small'' bodyparts as being scaled by $0.75$. We also group categories of limbs for environments with multiple appendages (i.e. humanoid torso includes the abdomen; humanoid thigh also includes the hips; all appendages encompass both the left/right or front/back simultaneously such that a modified thigh includes both thighs).

\subsubsection{Robot arm modifications}
In the OpenAI Striker and Pusher tasks, a 7 DoF arm tries to hit a ball into a hole or push a peg to a goal position respectively. We extend these tasks to randomly move the goal position for the Pusher task, and randomly move the ball start position for the Striker task. As in the original tasks, we bound the varied goal or start state within some restricted uniform distribution as domain appropriate. 

\subsubsection{Humanoid multitask}
We provide a humanoid multitask environment which combines the rewards for standing up and running in the same environment. The reward scale for this task is rather large, but aligns with the HumanoidStandup-v1 environment from OpenAI Gym. Additionally we provide a version of each environment with a sensor readout as in Section~\ref{sensor_envs}. When no wall is used, all sensors read zero. When a wall is used, each returns a distance to the wall as previously described.

\subsection{2D Navigation}
We also provide several novel 2D environments that focus on navigation tasks with continuous action spaces to enable benchmarking of learning tasks requiring an implicit memory. The tasks take place in a given occupancy grid map, similar to \cite{abbeel_vin}. We opt to make the layout
and shape of the obstacles as the only disambiguating feature for localizing within the map. Aside from that information, the environment does not have any texture mapping or other distinctive features. 




We provide three different types of navigation tasks, increasing in level of difficulty:
{\small\begin{itemize}
  \setlength\itemsep{0.05em}
 \item Image-based navigation where the agent has access to the entire map, including its own position within the map and the destination in the map as part of the image data.
\item State-based navigation, where the agent has access to its own position in the map and the distance and bearing to the closest obstacle. A simpler version also contains the destination coordinates.
\item Navigation based only on local range-and-bearing data around the agent using ray-tracing. It has to perform mapping and estimate its own position within the map (i.e. perform SLAM), while at the same time exploring to find the goal location, and learning to avoid obstacles. We also modify this with a simpler version, where the goal and current position are known as well.
\end{itemize}}
\vspace{-5pt}
We provide a reward of -1 for every timestep, -5 for obstacle collisions, and +10 for reaching the goal (which also ends the task, similarly to the MountainCar-v0 environment in OpenAI Gym). The action space is the bounded velocity to apply in the x and y directions.

\section{Multitask Sets} \label{task_sets}

We develop several sets of intuitive task groups which can serve as simple benchmarks which increase in complexity both within the group and in our listing order. The specific environment names can be found in Table~\ref{table:results_gravity}, ~\ref{table:results_body_parts}, ~\ref{table:results_sensor}, and ~\ref{table:results_arm}. For the navigation tasks, we list the environments inline here.

We introduce the following environment groups: 
\vspace{-5pt}
{\small\begin{itemize}
  \setlength\itemsep{0.05em}
    \item Modified environments with different gravity parameters\footnote{ \{BaseRunningEnv\} denotes one of the OpenAI Gym environments from: Humanoid, Hopper, Walker2d, HalfCheetah with \{GravityVariation\} from \{Half, ThreeQuarters, OneAndQuarter, OneAndHalf\}.} 
    \item Modified environments with sensor readouts (simply reading zero if no wall) and permuted with a random wall in the runner path
    \item The OpenAI Gym Striker environment with both random start position of the object as well as random goal state
    \item The OpenAI Gym Pusher environment with both random start position of the object as well as random goal state
    \item Learning to standup and run for a Humanoid model
    \item Learning to standup, run, and jump over walls for a Humanoid model
    \item Learning to run with different sized limbs with the base set of limbs encompassing \{Torso, Leg, Thigh, Foot\} and specific extra limbs listed below (i.e. example combinations look like: {\small HumanoidBigArm-v0, HopperSmallFoot-v0}).
    \item Learning to navigate and search in 2D environments using only current position and distance to closest obstacles ({\small State-Based-Navigation-2d-Map\{0-9\}-Goal\{0-2\}-v0})
    \item Learning to navigate and search in 2D environments observing current position, distance to closest obstacles, and known goal position ({\small State-Based-Navigation-2d-Map\{0-9\}-Goal\{0-2\}-KnownGoalPosition-v0})
    \item Learning to navigate and search in 2D environments observing only raytracing distance readouts ({\small Limited-Range-Based-Navigation-2d-Map\{0-9\}-Goal\{0-2\}-v0})
    \item Learning to navigate and search in 2D environments observing current position, raytracing distance readouts, and known goal position ({\small Limited-Range-Based-Navigation-2d-Map\{0-9\}-Goal\{0-2\}-KnownPositions-v0})
    \item Learning to navigate and search in 2D environments observing only the 2D map image with goal location and current position highlighted in different colors ({\small Image-Based-Navigation-2d-Map\{0-9\}-Goal\{0-2\}-v0}) 
\end{itemize}}

\section{Baseline Experiments}

We develop a basic experiment to run on the aforementioned groupings of the environments to demonstrate learning on a series of multiple similar tasks consecutively. We then evaluate the generalized performance across all of the environments using the final learned policy. For an initial baseline, we simply run the RLLab \cite{rllab} implementation of Trust Region Policy Optimization~\cite{TRPO} (TRPO) using an identical policy network to~\cite{qprop2016}\footnote{Size 100, 50, 25 hidden layers with rectified linear activations and a tanh output activation, and hyperparameters: step-size, $0.01$; GAE lambda, $1.0$; regularization coefficient, $1.0\cdot 10^{-5}$; number of epochs, $1000$; batch size, $50000$}. We train the same policy consecutively on each environment in a group in the same order as listed in Section~\ref{task_sets}. After having trained on a specific environment, we evaluate the current policy on that environment by running 20 sample rollouts. We then train on the next environment in the group, starting from that same policy. We finally evaluate the reward across 20 sample rollouts on each environment in a group using the final learned policy (which by then has been trained on every variation of the environment). While this isn't an explicit multitask learning approach, this provides basic insights (using a well-known reinforcement learning algorithm) into forward transfer and generalization of a policy on these task groupings.

Our baseline experiment results are found in Tables~\ref{table:results_gravity}, \ref{table:results_body_parts}, \ref{table:results_sensor}, \ref{table:results_arm}. In the case of modified Hopper tasks, modifying gravity and body part size has a profound effect on the system dynamics. As a result, we see that catastrophic forgetting \cite{catastrophic_forgetting_McCloskey} in the policy prevents generalization to earlier tasks. This can be seen in Tables~\ref{table:results_gravity} and~\ref{table:results_body_parts}. First, when evaluating the final policy on all of the previously trained environments (the ``Fully Trained'' column), performance decreases monotonically as we move backwards over the environments. Additionally, immediately after training on the earlier environments (the ``After Env Training'' column), the performance on the sample rollouts is much higher than that of the final policy (which has seen all the environments). This indicates that this group of environments are good indicators for demonstrating and overcoming catastrophic forgetting in multitask learning.

In other environment variations (modified HalfCheetah and Walker2d environments), the agent's final policy outperforms both training from scratch (as in Table~\ref{table:results_gravity}) and the `After Env Training'' result (as in Tables~\ref{table:results_gravity} and~\ref{table:results_body_parts}), which is evidence of positive forward transfer. The dynamics of these environment are not significantly perturbed by changes in physics, as the models have inherent stability. There remains significant room for future improvement upon our baseline. Future methods may achieve more efficient forward transfer between sequential environments. Furthermore, generalization across multiple tasks may come at a cost of higher variance in the policy (e.g. in Walker2d environments here). Future improvements may also focus on generalization with constrained variance across trials (and thus higher safety when learning on new environments).


For the Humanoid-exclusive variations and Wall variations (as in Tables \ref{table:results_body_parts} and \ref{table:results_sensor}), TRPO is not able to learn a policy which can jump over a wall or learn a good policy on Humanoid tasks in the small number of iterations which we ran (1000 iterations). The results we see are on a comparable scale to~\cite{rllab}.


In our new map navigation tasks, rewards remain at -1000, which is the initial lowest reward. That is, the agent never learns to find the goal using our default parameters and TRPO. This is to be expected as TRPO may not be suited for such a navigation task which requires large amounts of exploration with an extremely delayed reward. Other methods which encourage principled exploration and have a memory component to the policy may be more suitable for such tasks. We nevertheless share these environments with the community in an effort to drive investigation into creating complex policies for simultaneous localization, exploration and goal searching in settings where goals and obstacles vary between tasks.


\section{Conclusion}

Our initial release investigates adding flexibility to standard OpenAI gym MuJoCo environments: modifying gravity, adding sensor readouts and a random wall obstacle, perturbing body-part sizes, and adding random goal/start state positions for arm environments. We also add an original set of environments for learning policies in continuous navigation tasks. In future releases we also plan to add standard environments for: adding motor noise, arm environments where the end-goal position has a velocity (such that the arm must track the target), and making the sensor-based environments more realistic (and thus more transferable to real-world systems).





\bibliography{example_paper}
\bibliographystyle{icml2017}

\end{document}
